Automating the Design and Development of Gradient Descent Trained Expert System Networks

Jeremy Straub
Department of Computer Science
North Dakota State University

1320 Albrecht Blvd., Room 258
Fargo, ND 58108
p: +1 (701) 231-8196
f: +1 (701) 231-8255
e: jeremy.straub@ndsu.edu

**Abstract**

Prior work introduced a gradient descent trained expert system that conceptually combines the learning capabilities of neural networks with the understandability and defensible logic of an expert system. This system was shown to be able to learn patterns from data and to perform decision-making at levels rivaling those reported by neural network systems. The principal limitation of the approach, though, was the necessity for the manual development of a rule-fact network (which is then trained using backpropagation). This paper proposes a technique for overcoming this significant limitation, as compared to neural networks. Specifically, this paper proposes the use of larger and denser-than-application need rule-fact networks which are trained, pruned, manually reviewed and then re-trained for use. Multiple types of networks are evaluated under multiple operating conditions and these results are presented and assessed. Based on these individual experimental condition assessments, the proposed technique is evaluated. The data presented shows that error rates as low as 3.9% (mean, 1.2% median) can be obtained, demonstrating the efficacy of this technique for many applications.

## 1. Introduction

Artificial intelligence techniques find use in numerous applications ranging from making decisions regarding human applicants [1] to robotic command [2]. Problematically, many of these techniques cannot be readily understood by humans in order to verify that they are making decisions in the ways expected. In some cases, the decisions they make cannot be otherwise evaluated to ensure that they are valid or optimal. Perhaps most problematically, systems will work well in most cases, but fail in cases where certain key assumptions fail. Systems that are not human understandable or reviewable risk catastrophic failures, which could harm humans by causing them to be incorrectly assessed (when applying for credit or a job, for example) or putting them at risk of injury or death caused by an autonomously controlled robotic system.

In response to these issues, techniques that are explainable have been proposed. These eXplainable Artificial Intelligence (XAI) techniques [3] aim to change autonomous decision making from being an opaque process that is not understandable to one that is transparent to humans [4]. The techniques are responsive to fears and concerns [5] about AI operations. However, explainable techniques – in some cases – simply involve retrofitting existing systems [6] with other systems which try to explain how they work, demonstrating the system's process instead of justifying its results. These systems, prospectively, could obfuscate broken functionality or incorrectly explain why a correct answer was arrived at [7]. Other AI techniques have explainability capabilities included [6]; however, concerns about the performance of these systems have been raised with some [3] arguing the they cannot achieve the results of neural networks (though others [7] have argued that they suffer from asymmetric comparisons and insufficient development, compared to neural networks, to support these claims).

In [8], a machine learning trained expert system was proposed. This system begins with an application-specific rule-fact network and uses gradient descent techniques to optimize the network. It is not able to learn additional rule connections that could represent non-causal associations or other problematic or illegal relationships. This technique was shown to be able to rival the performance of neural networks under a variety of operating conditions. The efficacy of four training algorithms [8,9] was demonstrated and techniques for enhancing performance, via error reduction [10], were also demonstrated. While the system was readily trained using simulated [11] networks, which demonstrated the efficacy of manual network generation, the cost of actual network generation for real-world applications remains a key limitation of this technique, as compared to neural networks which learn their associations during the training process.

This paper proposes and evaluates a technique for responding to this limitation. Specifically, it evaluates the efficacy of an approach where a larger and denser-than-needed (for the application area) homogonous-structure network is developed and trained. Once initial training is performed, this network is then pruned to remove low-value association rules. For real-world applications, the simplified network would then be human evaluated and have rule (association) meanings assigned and evaluated. After these creation steps, the network is then trained again, using the process from [8]. The efficacy of this approach is compared to simulated manual generation techniques.

This paper continues with a discussion of relevant prior work (in Section 2). Section 3 presents the experimental system and the experimentation techniques used. In Section 4, the proposed approach is presented, in detail. Section 5 discusses and assesses the use of fully and densely connected networks. Next, Section 6 evaluates the use of an adaptive pruning technique. Section 7 compares training methods for use with this technique. Then, Section 8 discusses the manual processing steps required to ensure that the system doesn't create problematic non-causal, discriminatory or illegal linkages. Next, Section 9 compares the proposed technique to other approaches before concluding, in Section 10, and discussing planned and needed future work.

**2. Background**

This section reviews prior work in several areas that provides a foundation for the work presented herein. First, artificial intelligence, machine learning and the limitations of these techniques are discussed. Then, explainable AI is reviewed. Next, prior work on expert systems is considered. Following this, in Section 2.4, gradient descent techniques are explained. In Section 2.5, the prior work on gradient descent trained expert systems is presented. Finally, in Section 2.6, prior use of pruning with artificial intelligence techniques are reviewed.

*2.1. Artificial intelligence, machine learning and their limitations*

Artificial intelligence techniques come in many forms. Techniques based on insects [12], water flows [13] and approximating the functionality of the human brain [14] have been developed. Other techniques, such as particle swarm optimization [15] and expert systems [16], draw on logical foundations instead of nature analogs. A wide variety of techniques have been built by combining pre-existing ones. Particle swarm optimization, for example, has been used, in conjunction with neural networks, for training [17] and the Blackboard Architecture [18] is based on expert systems' foundations. In its many forms, artificial intelligence promises "extensive changes that will … affect all aspects of our society and life" [19]. It is poised to provide "unlimited, additional benefits that will open through the widespread usage of AI inventions" [19]. It is predicted that these benefits will be enjoyed in areas including "healthcare, education, the environment, and smart cities" [20]. Examples of AI benefit have already been identified in health care [21], fraud prevention [22], environmental protection [23], business [24] and marketing [25].

The work presented herein builds on the concepts of machine learning and rule-fact network expert systems. Multiple types of machine learning have been developed, previously. These include techniques, referred to as supervised learning [26], where inputs and target outputs are supplied, and those where the system is rewarded for producing desired behaviors [27]. An approach where the machine learning system simply identifies associations within data (without feedback) is also commonly used [28]. Expert systems utilize networks of rules and facts to make logical decisions based on combinations of data and rules that provide logical interconnections between these fact values. Expert systems are discussed further in Section 2.3.

For machine learning techniques which utilize feedback for training, a common approach used for optimization is called gradient descent [29]. This technique, which is discussed in more detail in Section 2.4, uses an iterative refinement process that modifies network weightings seeking to maximize the accuracy of system outputs.

Artificial intelligence techniques are not without their issues. For some application areas, there is a notable benefit to nefarious parties if they are able to interfere with or confuse the system. Adversarial neural network techniques [30], for example, have been demonstrated for attacking systems that use neural networks such as voice [31] and facial recognition [32] systems. AI's use in law enforcement and the judiciary creates other areas where an incentive may exist to attack its functionality. AI has been proposed for use in car crash investigation [33], legal text classification [34] and even for legal reasoning [35].

Even without an adversarial party, AI systems may fail to model actual causal relationships due to confounding variables and other issues. This may cause the system to operate accurately in cases where the confounding association holds true and catastrophically fail in other instances. Though not caused by adversary action, learning based on these associations may heighten systems' risk of successful attack by deception [36]. Issues of this type may also cause systems to run afoul of anti-discrimination and other laws through making decisions based on protected characteristics that sometimes correlate with predicted outcomes. Hallevy [37] and Hildebrandt [38] proffer that these types of issues could potentially cause individuals and firms to face criminal sanctions.

*2.2. Explainable AI*

Machine learning techniques have been demonstrated to be highly effective; however, many are inherently opaque [3]. In response to this issue and to allow human understanding and oversight of machine learning systems, a variety of XAI techniques have been proposed. Examples of XAI techniques include "Bayesian teaching" [39], "knowledge-to-information translation" [40], "evolutionary fuzzy" methods [41], "fuzzy relations and properties" [42] and "Shapley-Lorenz decomposition" [43]. Arrieta, et al. [6] proffer that XAI techniques can be grouped into two categories: those which inherently have "some degree of transparency" and, thus, are "interpretable to an extent by themselves" [6] and techniques which are designed to explain non-inherently explainable AI techniques [6].

Vilone and Longo [44] presented a XAI field taxonomy which distinguishes between techniques in terms of whether they are "model agnostic" or "model-specific". Ribeiro, Singh and Guestrin's [45] use of "anchor" propositions (statements highly associated with a particular result and dominant system behaviors) is an example of a model agnostic approach, while Setiono and Liu [46] and Palade, Neagu and Patton's [47] techniques for generating rules from neural networks are both examples of model-specific approaches.

XAI techniques have found use areas including recommendation systems [48], cyber intrusion detection [49] and "small-unit tactical behavior" planning [50]. They have also been employed for sales [48], lending [48] and fraud detection [48].

The techniques for these and other applications have been developed using a variety of common forms of artificial intelligence. For example, multiple fuzzy logic-based XAI techniques have been proposed, including those that use [51,52] and automatically generate [53] fuzzy rules and those which extract rules from neural networks [54]. Approaches which are "neural-symbolic" and combine "connectionist learning and sound symbolic reasoning" have also been discussed [55,56] as have approaches which extract rules from support vector machines [57] and "multi layer perceptrons" [47,58]. Systems [59] which use outputs from an opaque system as the inputs to a rule-based explainable system, making them the decision rationale, and which use diagramming [60,61] and simplified language [62] to explain rule-based system behaviors have also been proposed.

Despite these successes, current XAI capabilities are not sufficient for many applications. Additionally and problematically, many of the XAI techniques only seek to explain prior decisions that have already been made; thus, they don't constrain the system's operations. This means that while the XAI technique can help users understand or explain operations, it is not able to guarantee that the system will make decisions which are in line with prior explanations and assumptions or key criteria which is applicable in a particular instance.

## 2.3. Expert Systems

Rule-fact expert systems make logical inferences and deductions based on a collection of data elements (facts) and associations between them (rules) [63]. These systems were initially introduced in the 1960s and 1970s with two seminal systems, Dendral [64] and Mycin [65]. Expert systems have been used in numerous application areas. Examples of these include facial feature identification [66], medicine [67,68], education [69] and agriculture [70].

Multiple types of expert systems have been developed using different technologies. Systems that use fuzzy logic and fuzzy set concepts [71] as well as those that implement various optimization techniques [72] and hybrid systems [73], which use neural networks, have been designed or demonstrated.

Of particular note is a taxonomy for fuzzy logic-based expert systems developed by Mitra and Pal [74]. In this, they defined "fuzzy expert systems" as those which make use of use fuzzy sets (in conventional expert system designs) [74]. Another type, "fuzzy neural network" systems, are defined as systems which incorporate a neural network [74]. The related (to fuzzy neural networks) "connectionist expert system[s]" [74] also use neural networks. In this instance, the neural networks are used to generate rules. With "neuro-fuzzy expert systems," the system's knowledge base is also stored in the neural network.

The type of expert system that appears to be the closest to the gradient descent trained expert systems, presented in [8], is the type referred to by Mitra and Pal as a "knowledge-based connectionist expert system" [74]. This system was conceptually proposed to start with "crude rules" which would then be put into a neural network that would be used to produce optimized rules. While Mitra and Pal provided a brief conceptual system description, it does not appear that they fully designed or implemented this system type [8]. It also appears likely that the system could have suffered from the ability of the neural network to completely change the initial rules and by optimizing the node associations to a dissimilar configuration which may include illogical or non-causal relationships.

Diao, et al. [75] recently proposed a "convolutional rule interface network" system which uses a "belief rule base" to enhance the ability to interpret neural networks. This work, which focuses on classification, is similar to Mitra and Pal's in that it utilizes a neural network for rule processing. Diao's system features a "rule extraction layer" which leads to an "output layer". The work is also similar to the gradient descent trained expert system [8] in not using an activation function between nodes.

### *2.4. Gradient Descent*

Gradient descent techniques, such as back-propagation, have been used frequently for the optimization of neural networks. Given this, they are a logical choice for optimizing a system that conceptually combines neural network and expert system principles.

Gradient descent techniques are typically used to optimize weightings applied to the links between neural network nodes by using system inputs and desired outputs. The system is run, and a portion of the error between the network-produced output and the optimal output is applied throughout the network. Backpropagation [76], in particular, uses an iterative process for applying this error (starting from the output nodes and working backwards towards the input nodes).

As a relatively mature technique, a variety of approaches have been developed, including those that have been shown to enhance performance speed [77], techniques that attempt to increase accuracy by introducing noise [78] and some that seek to create attack resilience [79,80]. Other techniques have incorporated federated learning and momentum [81] and used evolutionary algorithms [82], speculative approaches [83] and spiking neural network concepts [84,85]. Yet other techniques have focused on supporting deep networks [86], memory use optimization [87], bias factors [88,89] and initial condition sensitivity [90]. A recent technique, proposed by Zhang, et al. [91], utilizes a combination of expert strategies and gradient descent for optimization.

Back-propagation is not the only technique that has been used for training neural networks. In particular, Kim and Ko [92] and Ma, Lewis and Kleijn [93] proposed techniques for neural network training which specifically avoid backpropagation. Others have used artificial intelligence techniques, such as genetic algorithms [94], Levenberg-Marquardt training algorithms [95], particle swarm optimization [96] and simulated annealing [96], for neural network training.

## 2.5. Gradient Descent Trained Expert Systems

A machine learning trained expert system approach was proposed in [8] which merges, conceptually, rule-fact network expert systems and neural networks. The technique utilizes a gradient descent machine learning technique directly on an expert system's rule-fact network. This has limited similarity to using neural networks to generate expert system rules, as proposed in [74]; however, because the technique performs the training on the existing expert system network, the pathways themselves cannot not change. The system can only optimize the weightings applied to network segments. This was shown, in [8], to facilitate training (for cases where the network under training is similar to the original network). It also prevents the system from learning non-causal (and potentially unethical or illegal) pathways (which would require adding a network segment).

The system proposed in [8] and used for this work includes an algorithm which distributes correction for the difference between the system-computed and ideal outputs to rules' fact weightings. This algorithm runs in a module which operates mostly independently of the expert system engine (though the engine is required to calculate the network's output for training). The amount of error correction that is applied to each rule is calculated based on the contribution of the rule to the run's target node's final value.

The concepts of partial membership and ambiguity are used. These create the potential for optimization and also provide reasoning capabilities beyond those of a basic Boolean rule-fact network-based expert system. The rule-fact network's facts can store values between 0 and 1 (instead of just true or false). Rules, then, instead of just being logical 'AND' operators, have weighting values which indicate how much contribution of each of the rule's two input facts has towards the output fact. Rules were designed to have just two input facts as the transitivity property of multiplication allows rules with more than two inputs to be built by combining multiple rules. Each rule's weighting ($W_1$ and $W_2$) values must be between 0 and 1 and $W_1 + W_2 = 1$.

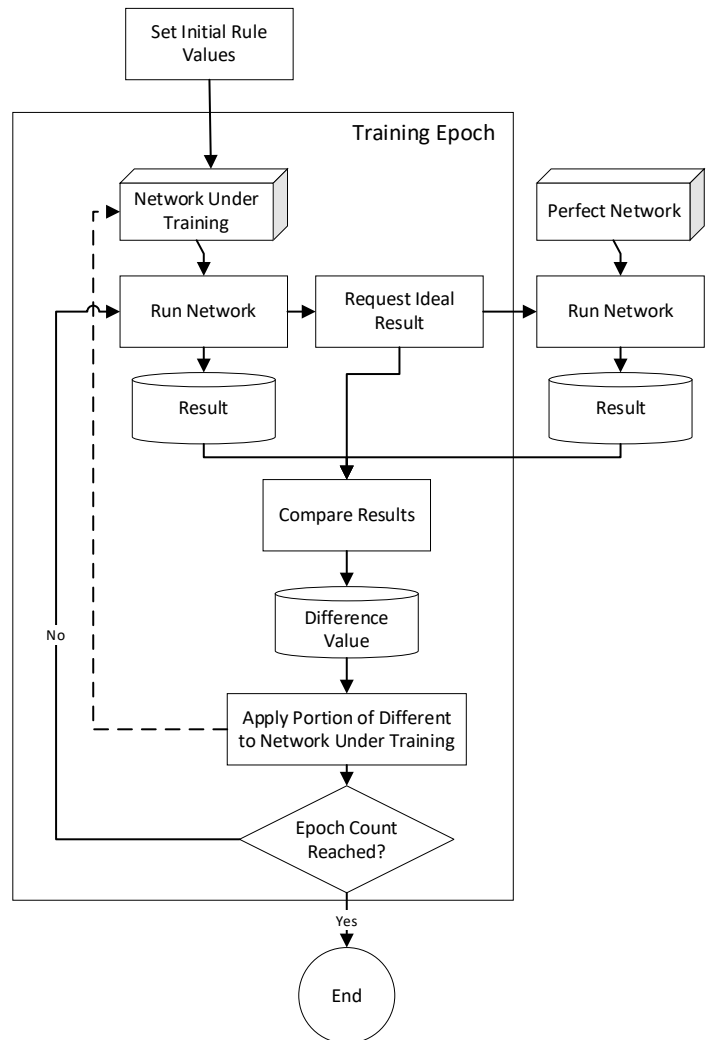

Figure 1. System Training Process [8].

An algorithm was developed [8] for training expert systems using gradient descent, as the networks used by expert systems differ significantly from those typically used for neural networks making the liner algebra typically used for optimizing neural networks inappropriate for use with expert system networks. The algorithm identifies rules that directly or indirectly contribute to the system's result value: first, contributing rules are identified and then, the system-under-training and 'perfect' system are run and a portion of the error between the systems' outputs is applied to the identified contributing rules. The amount of change applied is based on rules' contribution level and a velocity setting. The approach is depicted in Figures 1 and 2.

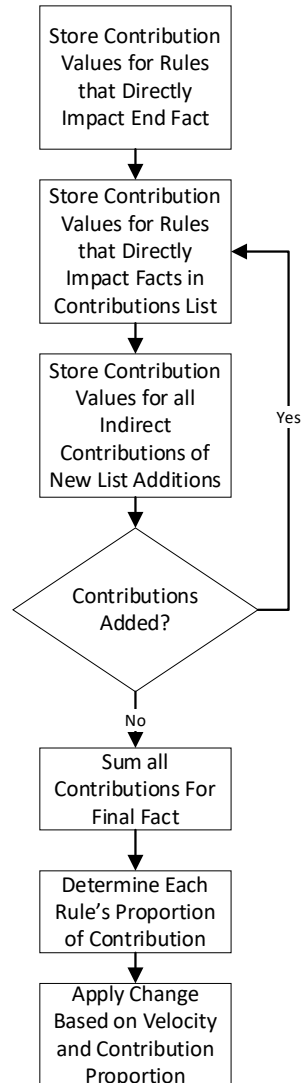

Figure 2. Algorithm for determining node change levels [8].

The training process, shown in Figure 1 – as used in [8] and described in [11], begins with a rule-fact network with the same structure as the perfect network; however, its initial rule weighting values are independent from perfect network's rule values. This same concept is used in this study; however, for the work presented herein, the initial network does not approximate the target network.

During the training process, both networks are run to generate results for a given target fact which are used to calculate the difference value for back propagation. This difference value is applied to the previously identified contributing rules for the target fact using the process shown in Figure 2. This training process is repeated the number of iterations specified by the training epochs parameter.

The algorithm begins by identifying all directly output-impacting nodes and then it identifies all nodes that impact the identified nodes, iteratively. It calculates their direct or indirect contribution level and adds them to the contributions list. This process is repeated until an iteration completes with no nodes being added. The difference applied to each rule weighting is calculated by dividing the contribution of the rule and by the total contribution of all rules and multiplying it by the velocity parameter and the normalized difference between the expected and actual value for the training run. The equation used to determine the contribution level of a rule, Ci, is [8]:

$$C_i = W_i \times \prod_{\{APT\}} W_{R(m,h)}$$

(1)

where $W_i$ is the weighting for rule, i, $W_{R(m,h)}$ is the weighting for a given rule in set APT (m and h indicate the rule and the particular applicable weight value, respectively) and {APT} is a set of all of the rules that a given rule passes through for its indirect contribution. Notably, while a rule may be a member of several rule-fact chains leading to a target fact and could have multiple contribution values, only the largest contribution value is retained and used contribution calculations. Once the rule contributions are identified, the difference value for a particular rule weighting, Di, is calculated using the equation [10]:

$$D_i = RC \times V \times DV$$

(2)

where RC is the rule contribution, V is the velocity value (which should be between 0 and 1) and DV is the difference value for the training run. This value is used to increase and reduce the weighting values assigned to the rule, based the comparative values of the rule's inputs. The RC value, for this equation, is calculated using the equation [10]:

$$RC = \frac{C_i}{\sum_{\{AC\}} C_i}$$

(3)

where $C_i$ is the contribution of a particular rule, i, and {AC} is the set containing all contributing rule nodes. Finally, the difference value for the run is calculated using the equation [10]:

$$DV = \frac{|R_P - R_T|}{MAX(R_P, R_T)}$$

(4)

where $R_P$ and $R_T$ are the results from the perfect and training networks, respectively, and MAX is a function which returns the largest value supplied to it as an input.

To define experimental runs, [8] used a number of parameters including the velocity value, the size and connectivity of the network, the number of training epochs and the type of training used. Two of the four forms of training presented in [8] are used in this study and discussed in Sections 2.5.1 and 2.5.2. Like in [8], the system is evaluated by assessing its efficacy at determining the value of a target fact in the rule-fact network. Also, similarly, networks are randomly generated for each testing iteration and the initial and final facts are randomly selected for each run. The initial fact is initialized at run commencement and the final fact's value is recorded at run completion and compared to the ideal system's value.

*2.5.1. Train Path – Same Facts*

The train path – same facts technique is the most basic form of training and was used throughout [8]. It utilizes a single set of facts and trains using a single path through the network. Start and end facts are randomly identified to define this training path. This approach will typically be the lowest cost to implement for real-world applications due to its minimization of data collection requirements.

*2.5.2. Train Path – Random Facts*

The train path – random facts training approach simulates data that has fact values that vary significantly. This approach, like with the previously discussed technique, uses a single path. It uses random fact values which are assigned for each training run. While the facts are different for each run, the ideal system's rule weightings do not change, and the system-under-training's rule weightings persist between training epochs and are updated each epoch (using the error correction distribution method discussed previously). This approach, thus, allows the network-under-training to be exposed to more conditions and nuanced data. In [8], the approach was shown to produce slightly less average error than the train path – same facts technique. Notably, the assumption that this simulation technique is based on (significantly different fact values) will not be true for many applications. Thus, for instances where fact values are similar or have minimal variation, the train path – same facts training technique is more representative.

## 2.6. Pruning and AI Techniques

The concept of pruning has been applied to a variety of artificial intelligence techniques. Prior work demonstrated its efficacy for enhancing Blackboard Architecture networks (which are logically similar to rule-fact expert system networks) [97,98].

Several examples of its use with neural networks also exist. Reed [99] discussed the use of techniques such as sensitivity calculation, penalty application, weight decay, interactive pruning, bottleneck considerations and the use of genetic algorithms. Blalock, et al. [100] reviewed 81 papers which utilized pruning to reduce neural network size and enhance performance; however, they note that the papers' inconsistencies resulted in difficulties comparing the findings between them. A shorter survey of 45 papers by Xu, et al. [101] focused on the techniques that could be used and the trade-offs that system designers who used pruning faced.

Techniques based on "input importance" [102], probability [103] and "relevance scores" [104], in particular, are relevant to the techniques discussed herein. Wang, et al. [105] also discuss the related approach of identifying the subset of a neural network which produces the bulk of its capability, allowing the rest to be pruned with limited impact. Augasta and Kathirvalavakumar [106] demonstrated the efficacy (and analyzed the comparative performance) of several different pruning techniques on multiple datasets. While they showed that some techniques outperform others, for given datasets, they also demonstrated the key tradeoff between pruning speed, network operation speed and accuracy.

## 3. Experimental system and experimentation

This paper builds on the prior work presented in [8] and uses a similar assessment methodology to it (which was discussed, in more detail, in [11]). The experimental system utilized is based on and enhances the system used in [8] and is discussed in Section 3.1. The design of the rule-fact networks utilized for experimentation is discussed in Section 3.2. Then, experimental procedures are discussed in Section 3.3.

*3.1. Experimental System Design*

The data collection presented in subsequent sections of this paper was collected from the operations of a modified version of the rule-fact expert system software presented in [8]. This system includes an expert system engine which processes rules in a forward fashion and a training module, which operates mostly independently of the expert system engine (it is used as in the training process for determining the rule-fact networks' outputs), that optimizes the network's rules' weightings.

As with the previous work, the expert system utilized supports the partial membership and ambiguity, (allowing facts to have a probabilistic or partial membership value between 0 and 1 as opposed to just a value of true or false). Due to this, rules have weighting values for the comparative impact of both input facts on the value of the output fact. These weightings must be between 0 and 1 and their sum must be 1. This is distinctly different from classical rule-fact expert systems which might simply have rules that require two (or more) facts be true, as a precondition, and assert a third fact, if the precondition is met.

The system used for [8] was extended to support the pruning and filtering techniques which are described in Section 4. The data collected for each experimental run was the same as was used in [8].

### 3.2. Experimental Rule-Fact Networks Design

For this experimentation, the perfect (training input) network is created and then another network is created which is trained using the inputs and outputs of the perfect network as inputs to its training process. The perfect network is used as the basis for the evaluation of the efficacy of the algorithm being tested. It serves as a perfectly modeled reality (which can be perturbed, if needed, to create experimental conditions with known discrepancy and noise levels). The network under testing is created and trained and its output is compared to the output from this model, as described in [11].

In addition to the training algorithm, most experimental conditions also include a pruning step (some do not, for purposes of comparison), where nodes are removed from the network to attempt to produce a network that functionally parallels the perfect network without the benefit of knowing the configuration of the perfect network. This is distinctly different than what was investigated in [8], where knowledge of the perfect network was a key input.

Each experimental condition is defined in terms of the number of rules and facts in the network, the training algorithm that is used and the pruning algorithm and options selected. All of these are user-defined simulation parameters.

The perfect network is generated based on user-supplied parameters. Rules are created and facts are selected at random as input and output facts for these rules. The rules are checked against the existing ruleset to prevent duplication.

Several forms of networks-under-training were utilized. Layered networks are highly connected (every fact node in each layer is connected, via rules, to all nodes in adjacent layers) and mirror neural networks in configuration. Densely connected and random networks are made with random connections between facts (via rules) to, in the case of densely connected networks, a given level of connectedness. Random networks have a given number of rules that are placed in between facts (randomly). Fully connected networks interconnect each fact with every other fact (also using rules). In the two instances where data for perfect networks-under-training is presented (for comparison purposes), these networks are generated by copying the rule-fact structure of the network but resetting the rule weightings.

### 3.3. Experimental Procedures

During each experimental run, the expert system is evaluated in terms of its ability to determine the value of a selected target final fact. This target fact is selected from a rule-fact network which is created randomly for each system run.

For each run, initial and final facts are randomly selected from the network. The initial fact is set to an initial value of 0.99. The final fact's value is recorded at the end of the run, for evaluation purposes. A run continues until no facts are modified during an iteration of system operations. Runs are immediately terminated in cases where the final fact's initial value satisfies run completion criteria.

To compare different pruning approaches under various training approaches and other experimental conditions, the performance of the system was evaluated 1,000 times for each experimental condition. Evaluation runs that did not complete (i.e., because of network connectivity and other random generated network issues) and runs that complete immediately were excluded from the calculations. To facilitate comparison, all of the data collection runs were performed on a collection of 20 workstations with the same hardware configuration.

## 4. Techniques and their implementation in the gradient descent trained expert system

The concept explored in this paper is to automate the creation of gradient descent trained expert systems by starting with a larger and denser network and pruning it down to the essential elements. An example of pruning is shown in Figure 3. The pruning process has two potential applications: refining human-created networks and automating network creation.

For human created networks, pruning could be used to refine the network and reduce processing time by removing rules and facts that are not used, infrequently used, or even problematic. Thus, in a best case, it might help to mitigate a misunderstanding about how a phenomenon works, via identifying an error in the model. In other cases, it might expedite processing and simplify the model – making it easier for humans to understand. Any rules or facts suggested for removal by the pruning process could be assessed by human reviewers to validate the appropriateness of their removal. Of course, it would be critical to ensure that representative data is used for the punning process and that removed rules and facts don't handle exception cases where their removal could result in significant errors. When used for this application, assessment of both the average error and the identification of large single case changes would be needed to ensure that the changes that are made are both generally beneficial and not detrimental to specific cases.

While potentially beneficial, the use of pruning for the refinement of human created networks is left as an area for future work. This paper considers the efficacy of pruning for the automation of network creation.

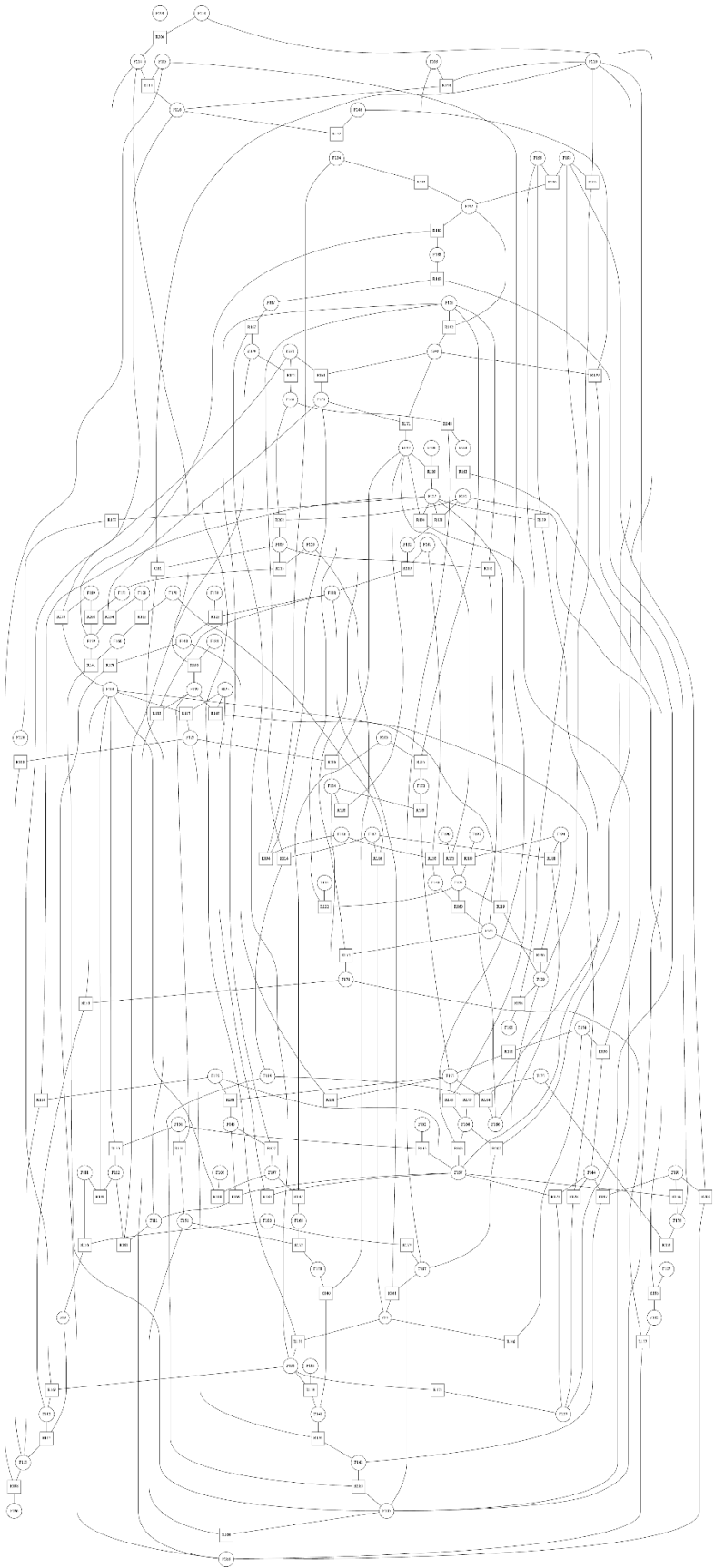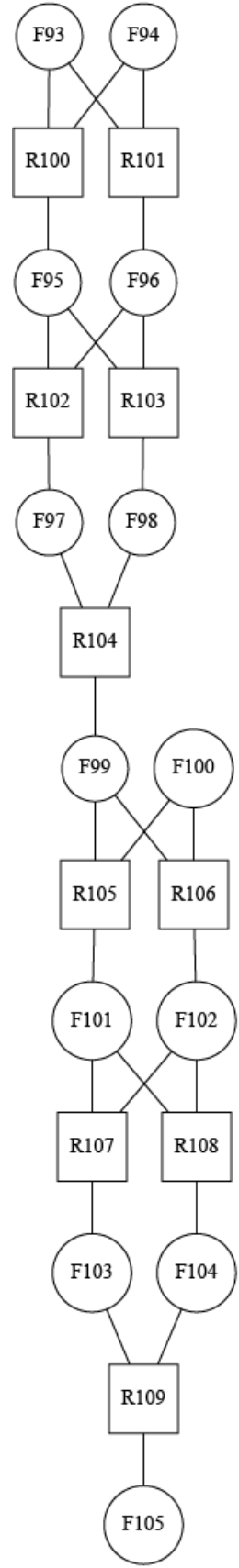

Figure 3. Rule-fact network: (a) prior to (left) and (b) after (right) pruning process.

*4.1. Pruning for Network Creation*

Pruning for network creation is premised on the ideal network being contained within a larger network. Pruning is used to remove the rules and facts that are not part of this ideal network, leaving only the ideal network remaining.

A fully connected network with all of the rules and facts of the ideal network (i.e., one that has all requisite inputs and outputs and sufficient intermediary rules and facts to handle the pathways within it) would be guaranteed to contain the ideal network. Problematically, though, fully connected networks are large and require significant time and resources to operate and prune. Networks which are not fully connected could, prospectively, still contain the ideal network. They may also contain networks which may sufficiently approximate the ideal network – making them suitable for practical use.

With the techniques presented herein, a network would be created (several different initial types of networks are evaluated in Section 5 and subsequent sections), trained and then pruned. The output network would be evaluated by a human subject matter expert who would identify internal facts and rules' meanings and look for problematic associations (such as those which might be oversimplifications, stereotypes, biases or otherwise non-causal). The human subject matter expert could then add (or add back, if they were pruned) any missing associations that they identify or remove any associations that they identify as problematic, illegal or non-causal. Once the pruning and human expert review were completed, the final network would be trained for optimization purposes and then would be ready for use. This process makes gradient descent trained exert system learning closer to neural network techniques, while maintaining the known meaning of the rules and facts (which are labeled by the human expert) and defensibility and explainability of the system and its decision-making.

*4.2. Contribution Pruning*

Contribution pruning looks at the impact of each rule and removes rules that make no contribution or make a contribution below a specified threshold to any fact or a specific target fact. Contributions are already known, as these are used for the training process, as described in Section 2.5. Equation 1 is used to calculate the contribution of each rule. Each rule's highest contribution to each fact is determined by the system – both in terms of direct and indirect impact. These values are compared to the specified threshold for pruning decision-making.

While the contribution levels would seem to provide a good heuristic for pruning, there are several issues with this approach. First, just because a contribution exists, this doesn't necessarily mean that it is a positive contribution. Some contributions may negatively impact network operations – and this impact may increase as other contributions are removed via pruning. Second, in some cases, networks may be comprised of numerous small impacts that add to a desirable result. Conceivably, the rules providing these small impacts could all be removed by pruning or a network could be unprunable at a threshold level that retains them.

Thus, while the contribution metric may be a useful metric for identifying potential nodes to prune, it isn't the best for making the final decision on whether to prune them or not. Assessing system performance with training data is best positioned to aid this decision making.

*4.3. Adaptive Pruning*

This is what adaptive pruning does, as shown in Figure 6. Rules are identified for pruning. They are then suspended – allowing the network to run without them temporarily – and the result is compared to the result without the suspension. If the performance declines by removing the rule, it is reinstated. However, if removing the rule has no impact or enhances performance, it is removed. Every rule in the network is tested to determine whether it makes a positive contribution to network performance or not.

If a network is of low quality, the pruning process can demonstrate this. A filter mechanism, called active filtering, is included to identify these low-quality networks. If a network's pruning process completes without any rules remaining (meaning that all were identified as unhelpful), the network can be dropped. Without filtering, the pre-pruning network is utilized. The use of this filter would depend on whether operational needs require a network (even a low quality one) to remain after training.

Once the pruning is complete, the network is optimized and is ready for use.

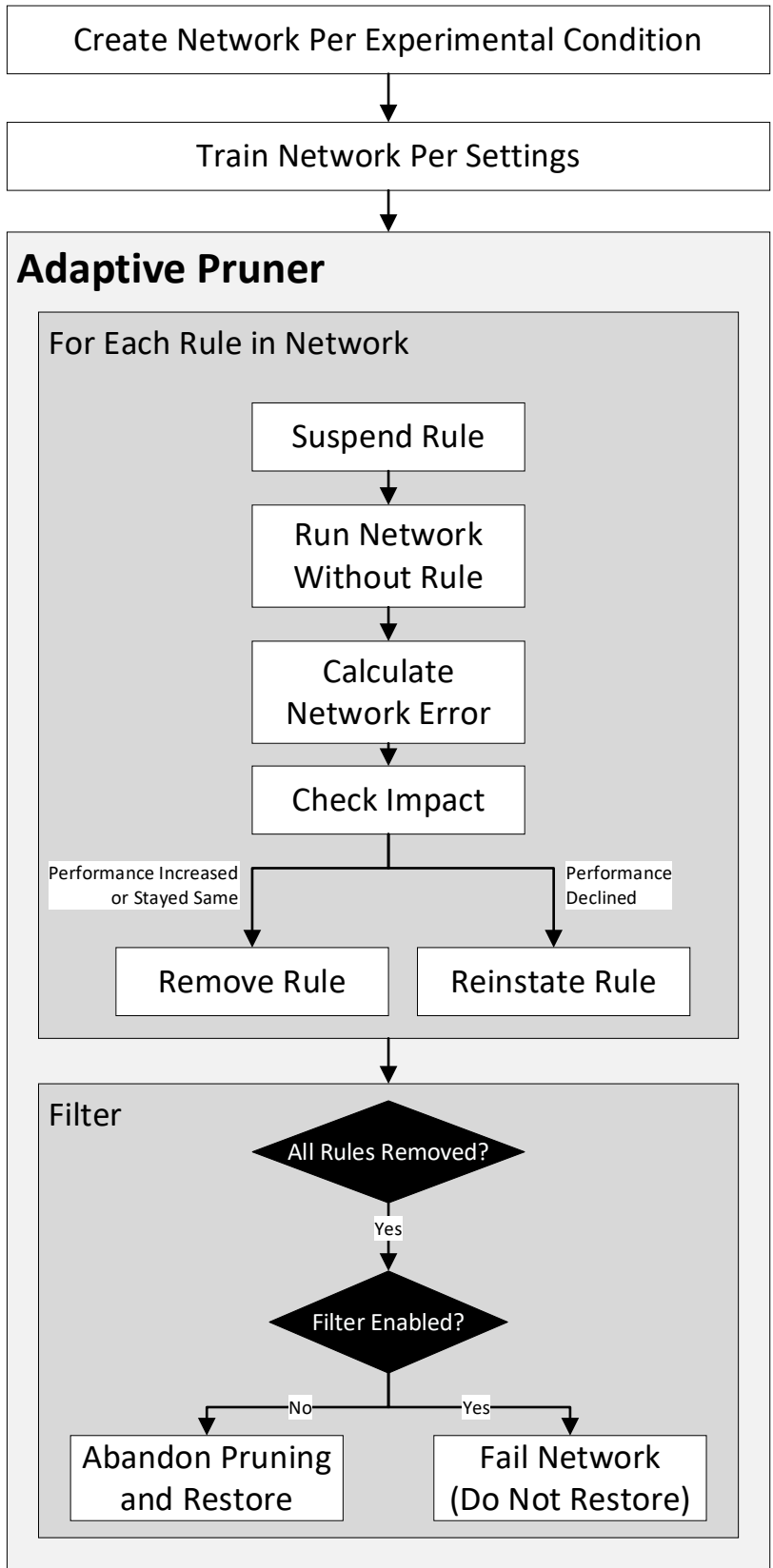

Figure 6. Adaptive pruning.

## 5. Fully and densely connected and layered networks

In addition to using network types from prior work (such as the perfect, random, fully connected and dense networks), this paper introduces a layered expert system network resembling the networks used by neural networks. As shown in Figure 7, each node in each layer of the network is connected to all nodes in the adjacent layers and no other nodes.

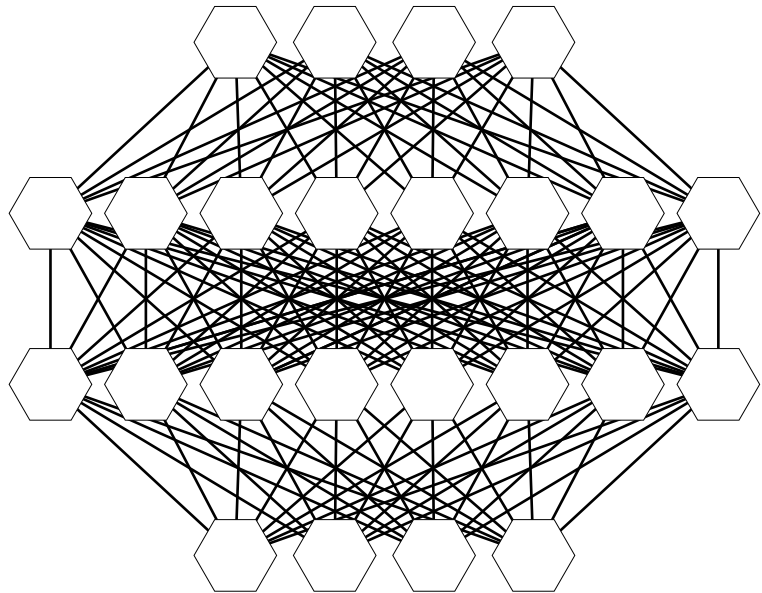

Figure 7. Layered Network (modified from [11]).

This section characterizes the performance of all of these network types, without pruning, to demonstrate the relative efficacy of trained layered networks (versus the perfect network and other network types) and to provide a baseline for comparing the pruned networks to.

### 5.1. Network types

This section characterizes the performance of the different network types. The performance of 10 rule / 10 fact networks with 100 training epochs is presented in Table 2 (the hyperparameters for this experiment are presented in Table 1). A layered network configuration that is five facts wide and five facts deep is presented for comparison. The table presents the mean and median of the error values. It also includes the average (mean) of the high-error (AV high) and low-error (AV low) networks and the number of the high-error (CT high) and low-error (CT low) networks.

As is to be expected, the perfect network type performs the best, with a mean error of 9.4% and median error of 3.9%. Notably, it produces the most low-error networks; however, other approaches produce more network completions (networks that are able to be trained and are not in an initially complete) overall. The dense and fully connected networks, albeit with 33.7% and 31.8% error, produce more total completions (but less low-error ones). There are 447 dense network completions and 435 fully connected network completions. Layered networks produce only about 40% of the low-error completions of the perfect networks. However, they produce a total of 410 completions, outperforming random networks (in terms of low-error and overall completions). Notably, layered networks also marginally outperform densely and fully connected networks in terms of mean and median error levels.

Table 1. Experimental hyperparameters.

| Model hyperparameters | | Experimental parameters | Algorithm hyperparameters | | |
|---|---|---|---|---|---|
| Facts | Rules | Network | Training epochs | Velocity | Training approach |
| 10 | 10 | Per condition | 100 | 0.1 | Base |

Table 2. Performance of Previous Network Types with 10 rules and facts and 100 training epochs.

| | Mean | Median | AV High | AV Low | Ct High | Ct Low |
|---|---|---|---|---|---|---|
| Perfect | 0.094454 | 0.03966 | 0.273003 | 0.030813 | 72 | 202 |
| Random | 0.285527 | 0.227727 | 0.391896 | 0.034801 | 132 | 56 |
| Fully Connected | 0.318374 | 0.270001 | 0.418252 | 0.037137 | 321 | 114 |
| Dense (50%) | 0.337321 | 0.296341 | 0.441898 | 0.038918 | 331 | 116 |
| Layered | 0.306912 | 0.253759 | 0.37114 | 0.041971 | 330 | 80 |

The results are similar for larger networks (fully and densely connected networks are not included in this analysis due to the extensive runtime required for larger networks). These results are presented in Table 4 and the experimental hyperparameters are presented in Table 3. The perfect networks outperform in terms of mean, median and low error network completions. Layered networks have a higher mean and median error than both the perfect and random networks. They do, though, produce the most overall completions (albeit mostly high error ones).

Table 3. Experimental hyperparameters.

| Model hyperparameters | | Experimental parameters | Algorithm hyperparameters | | |
|---|---|---|---|---|---|
| Facts | Rules | Network | Training epochs | Velocity | Training approach |
| 100 | 100 | Per condition | 100 | 0.1 | Base |

Table 4. Performance of Previous Network Types with 100 rules and facts and 100 training epochs.

| | Mean | Median | AV High | AV Low | Ct High | Ct Low |
|---|---|---|---|---|---|---|
| Perfect | 0.086363 | 0.052838 | 0.202517 | 0.038846 | 72 | 176 |
| Random | 0.226507 | 0.183325 | 0.314367 | 0.042799 | 92 | 44 |
| Layered | 0.321855 | 0.295944 | 0.36373 | 0.049668 | 351 | 54 |

## 5.2. Densely connected networks

Focus now turns to the impact of using different density levels for networks. Networks with 10 rules and 10 facts, different density levels, and 100 epochs of training are assessed in Table 6 (experimental hyperparameters are presented in Table 5).

Table 5. Experimental hyperparameters.

| Model hyperparameters | | Experimental parameters | Algorithm hyperparameters | | |
|---|---|---|---|---|---|
| Facts | Rules | Network | Training epochs | Velocity | Training approach |
| 10 | 10 | Densely connected – density level per condition | 100 | 0.1 | Base |

Table 6. Densely connected networks at different density levels with 10 rules and facts and 100 training epochs.

|  | Mean | Median | AV High | AV Low | Ct High | Ct Low |
|---|---|---|---|---|---|---|
| Density 10 | 0.279718 | 0.188512 | 0.403689 | 0.041544 | 317 | 165 |
| Density 20 | 0.274329 | 0.178582 | 0.392511 | 0.040883 | 320 | 162 |
| Density 30 | 0.295146 | 0.223763 | 0.429507 | 0.041355 | 306 | 162 |
| Density 40 | 0.310152 | 0.245 | 0.436385 | 0.032625 | 299 | 136 |
| Density 50 | 0.337321 | 0.296341 | 0.441898 | 0.038918 | 331 | 116 |
| Density 60 | 0.317607 | 0.258066 | 0.428069 | 0.039277 | 320 | 127 |
| Density 70 | 0.329111 | 0.275333 | 0.430254 | 0.037971 | 308 | 107 |
| Density 80 | 0.323739 | 0.284874 | 0.434888 | 0.035606 | 337 | 130 |
| Density 90 | 0.327556 | 0.2486 | 0.446413 | 0.036622 | 328 | 134 |

As is shown by the data, training densely connected networks, at the 100 epoch level, produces relatively consistent results across different density levels, with error increasing with increased density, up to density level 50, before declining slightly (and not increasing further after this point). The lowest mean error is produced by a density level of 20, while the highest was produced by a density level of 50. The lowest median was also produced by 20 density and the highest median was also produced by density 50. Notably, the number of low-error completions are the highest for density levels 10-30. The number of overall completions are also highest at the lowest density levels: density 10 and density 20 have 482 overall completions. The lowest level of total completions occurs at density 70, with density levels 80 and 90 reaching near the total completion level of the lowest density levels (467 and 462, respectively).

While there are demonstrable differences between density levels, none of the density levels is approaching the accuracy of the perfect network (though some outperform the layered network error levels presented in Tables 2 and 4). Given the better performance of the lowest density level (and thus simplest / lowest connectivity) networks, it may be that additional training, beyond 100 epochs, could be beneficial to further reducing error.

### 5.3. Layered networks

Now, the layered network configurations are tested in more detail. First, different combinations of layers (network depth) and nodes per layer (network width) are tested. Notably, while networks of different widths and depths are considered, all networks have 5 input and 5 output layer nodes. This data is presented in Table 8 and the relevant hyperparameters are presented in Table 7.

Next, different levels of training are evaluated for different network configurations. This data is presented in Tables 10 to 13 (and the relevant hyperparameters are presented in Table 9).

Table 7. Experimental hyperparameters.

| Model hyperparameters | | Experimental parameters | Algorithm hyperparameters | | |
|---|---|---|---|---|---|
| Facts | Rules | Network | Training epochs | Velocity | Training approach |
| Per condition | As required for connectivity | Layered – configuration per condition | 100 | 0.1 | Base |

Table 8. Performance of training layered networks with different layer depths and widths and 100 epochs of training.

|    |     | Mean     | Median   | AV High  | AV Low   | Ct High | Ct Low |
|----|-----|----------|----------|----------|----------|---------|--------|
| 5D | 5W  | 0.321855 | 0.295944 | 0.36373  | 0.049668 | 351     | 54     |
| 5D | 10W | 0.326607 | 0.284656 | 0.376782 | 0.051385 | 373     | 68     |
| 10D| 5W  | 0.335578 | 0.310119 | 0.388317 | 0.052412 | 349     | 65     |
| 10D| 10W | 0.328842 | 0.309666 | 0.381339 | 0.046392 | 382     | 71     |
| 5D | 15W | 0.335996 | 0.306505 | 0.392098 | 0.046796 | 366     | 71     |
| 15D| 5W  | 0.327027 | 0.286523 | 0.393626 | 0.048298 | 339     | 81     |
| 10D| 15W | 0.342526 | 0.330017 | 0.388482 | 0.053083 | 359     | 57     |

The performance, in terms of mean and median doesn't present any overt trends. However, the total completions and low error completions do show an interesting pattern. The middle complexity networks – 5x10 (and vice versa), 10x10 and 5x15 (and vice versa) – show the highest overall completions, outperforming the 5x5 and 10x15 networks.

The impact of training on the layered networks, as shown in Tables 10 to 13, appears to be somewhat limited. All of the different training levels appear to produce approximately the same mean error. For the 5x5 and 10x5 networks, the lowest mean error is recorded at the 10-epoch training level. The 5x10 network has its best mean and median error at 25 epochs. The 10x10 network has its best mean and median error at 50 epochs. The 5x5 network shows a limited median performance enhancement at 250 epochs. The 10x5 networks' best median is also at 250 epochs of training.

Table 9. Experimental hyperparameters.

| Model hyperparameters | | Experimental parameters | Algorithm hyperparameters | | |
|---|---|---|---|---|---|
| Facts | Rules | Network | Training epochs | Velocity | Training approach |
| Per condition | As required for connectivity | Layered – Config per condition | Per condition | 0.1 | Base |

Table 10. Performance of training layered networks with 5 depth and width and different training levels.

|            | Mean     | Median   | AV High  | AV Low   | Ct High | Ct Low |
|------------|----------|----------|----------|----------|---------|--------|
| 10 Epochs  | 0.317052 | 0.280859 | 0.379867 | 0.046162 | 345     | 80     |
| 25 Epochs  | 0.317059 | 0.292064 | 0.37769  | 0.049398 | 362     | 82     |
| 50 Epochs  | 0.333393 | 0.307487 | 0.384336 | 0.05102  | 388     | 70     |
| 100 Epochs | 0.321855 | 0.295944 | 0.36373  | 0.049668 | 351     | 54     |
| 250 Epochs | 0.34122  | 0.253859 | 0.4081   | 0.051138 | 360     | 83     |

Table 11. Performance of training layered networks with 5 depth and 10 width and different training levels.

|            | Mean     | Median   | AV High  | AV Low   | Ct High | Ct Low |
|------------|----------|----------|----------|----------|---------|--------|
| 10 Epochs  | 0.329129 | 0.294074 | 0.379779 | 0.048919 | 343     | 62     |
| 25 Epochs  | 0.299744 | 0.274681 | 0.361781 | 0.047906 | 341     | 84     |
| 50 Epochs  | 0.31031  | 0.294488 | 0.378507 | 0.048017 | 350     | 91     |
| 100 Epochs | 0.326607 | 0.284656 | 0.376782 | 0.051385 | 373     | 68     |

|  | 250 Epochs | 0.316242 | 0.275931 | 0.382547 | 0.051022 | 364 | 91 |

Table 12. Performance of training layered networks with 10 depth and 5 width and different training levels.

|  | Mean | Median | AV High | AV Low | Ct High | Ct Low |
|---|---|---|---|---|---|---|
| 10 Epochs | 0.3187 | 0.284321 | 0.380831 | 0.048288 | 383 | 88 |
| 25 Epochs | 0.333189 | 0.303835 | 0.392376 | 0.058231 | 367 | 79 |
| 50 Epochs | 0.325838 | 0.283239 | 0.379449 | 0.052572 | 367 | 72 |
| 100 Epochs | 0.335578 | 0.310119 | 0.388317 | 0.052412 | 349 | 65 |
| 250 Epochs | 0.323675 | 0.283097 | 0.377988 | 0.045991 | 363 | 71 |

Table 13. Performance of training layered networks with 10 depth and 10 width and different training levels.

|  | Mean | Median | AV High | AV Low | Ct High | Ct Low |
|---|---|---|---|---|---|---|
| 10 Epochs | 0.319775 | 0.280015 | 0.386236 | 0.04745 | 336 | 82 |
| 25 Epochs | 0.326446 | 0.284343 | 0.380999 | 0.053679 | 345 | 69 |
| 50 Epochs | 0.315842 | 0.294875 | 0.374025 | 0.039082 | 352 | 74 |
| 100 Epochs | 0.328842 | 0.309666 | 0.381339 | 0.046392 | 382 | 71 |
| 250 Epochs | 0.343437 | 0.32454 | 0.391424 | 0.052143 | 346 | 57 |

The networks also have their best performance, in terms of total completions and low-error completions at different levels. The 5x5 networks have their best total completions at 25 epochs and their most low-error completions at 250 epochs. The 5x10 networks have their most low-error completions at both 50 and 250 epochs and their most overall completions at 250 epochs. The 10x5 networks, conversely, perform the best for both low-error and overall completions at 10-epochs of training. Finally, the 10x10 networks have their most low-error completions at 10 epochs and their most total completions at 100 epochs.

From the foregoing, there seems to be little correlation between significant amounts of additional training and enhanced accuracy, for the layered networks. Given the similarity of the layered networks to neural networks, this data would tend to suggest that the gradient descent expert system training algorithm doesn't work well with this type of network in its base form; however, pruning inherently restructures the network and is shown to resolve this limitation.

*5.4. Limitations of fully and densely connected and layered networks*

Fully and densely connected networks, of a sufficient size, are guaranteed (in the case of fully connected networks) or likely (in the case of densely connected networks) to contain the ideal network for a phenomenon. However, they rapidly grow in network size, in terms of the number of interconnecting rules required, as the number of facts increases. Because of this, they become slow to train and unwieldly to operate. The key reasons that only small network sizes are used for analysis of fully connected networks (in [8] and herein) are the slowness and resource requirements of larger networks.

Additionally, while a fully connected network must contain the relevant ideal network, a combination of training and pruning decisions may result in the ideal network not emerging. Instead, a less-than-optimal network may emerge as the best performing. Densely connected networks trade time and resource level reductions (commensurate with the level of density) for a reduced search space that is, thus, less likely to contain the true ideal network. Again, whether the densely connected network does, in-fact, contain the ideal network or not, an alternate network may emerge as the best performing through the training and pruning processes.

Because of these factors, the fully and densely connected networks are unlikely to be appropriate for many applications.

Layered networks, on the other hand, have found significant use in modeling real-world phenomena in neural networks. They inherently simplify the potential connections between nodes – limiting nodes to only communicating with neighbors on adjacent layers (and, initially, communicating with all nodes on the adjacent layers). Like densely connected networks, layered networks trade off reduced complexity for potentially excluding the ideal network from being available for identification. Like fully and densely-connected networks, though, the most ideal network present may not emerge, due to training and pruning decisions. Layered networks, in particular, are inherently limited in their ability to model networks that have loop structures present in them or other rule links that are not part of a one-way data flow model. Additionally, the number of layers inherently limits the length of a solution rule-fact network chain in a layered network. This requires larger (deeper) networks with more facts to model some phenomena with long-chain characteristics.

Layered networks are faster and easier to work with than fully and densely connected ones, due to the significant reduction in the number of interconnecting rules required, because of the layered configuration. Their demonstrated efficacy for many phenomena, as shown by numerous neural network implementations, makes them of considerable interest for evaluation and they are, thus, considered extensively throughout the rest of this document.

## 6. Adaptive pruning

This section presents data to characterize the performance of the adaptive pruning technique, which was presented in Section 4.3. Table 15 characterizes the performance of the adaptive pruner on layered networks of different sizes and configurations (relevant hyperparameters are shown in Table 14). Notably, the adaptive pruning techniques improve over the approximately 30%-35% mean error and 25%-30% median error levels of the contribution-based pruning techniques presented in the previous section. The 5x10 networks perform the most poorly, with slightly higher error levels than the 5x5 and 10x5 networks.

Table 14. Experimental hyperparameters.

| Model hyperparameters | | Experimental parameters | Algorithm hyperparameters | | |
|---|---|---|---|---|---|
| Facts | Rules | Network | Training epochs | Velocity | Training approach |
| Per condition | As required for connectivity | Layered – Config per condition  Pruning Type – Adaptive  Prune Epoch – 20 | 100 | 0.1 | Base |

Table 15. Performance of adaptive pruning with networks of different depth and width.

| | | Mean | Median | AV High | AV Low | Ct High | Ct Low |
|---|---|---|---|---|---|---|---|
| 5D | 5W | 0.260715 | 0.212531 | 0.417192 | 0.029203 | 253 | 171 |
| 5D | 10W | 0.282754 | 0.230784 | 0.456991 | 0.020872 | 248 | 165 |
| 10D | 5W | 0.263405 | 0.218168 | 0.410623 | 0.024966 | 264 | 163 |

Tables 17 to 20 characterize the performance of adaptive pruning for different levels of training and different prune points for two different network configurations (5x5 and 5x10). Relevant hyperparameters for these experiments are presented in Table 16. The data in Tables 17 and 18 compares different training levels with pruning performed 5 epochs before training completion (allowing for 5 epochs of post-pruning training). While some variation is shown (and a downward trend in error with higher training levels is present in the Tables 17 and 18 data), the difference in performance is limited and error levels don't reach a point where the system would be useful for many applications. Notably, the low-error average is in the 2-3% range for both, meaning that the low-error networks would be well suited for many applications. Additionally, as the average of the low-error networks is lower than for the contribution-based techniques, the adaptive pruning techniques do not appear to be packing the higher end of the low-error range to generate the higher low-error count numbers. The combination of these two values (higher low-error completions and lower low-error average error) demonstrates the efficacy of the technique. However, due to the presence of numerous high-error networks, its overall performance still falls short of what is required by many applications.

Table 16. Experimental hyperparameters.

| Model hyperparameters | | Experimental parameters | Algorithm hyperparameters | | |
|---|---|---|---|---|---|
| Facts | Rules | Network | Training epochs | Velocity | Training approach |
| Per condition | As required for connectivity | Layered – Config per condition  Pruning Type – Adaptive  Prune Epoch – Per Condition | Per condition | 0.1 | Base |

Table 17. Performance of adaptive pruning with different levels of training and pruning points for 5 deep, 5 wide layered networks.

| Train | Prune | Mean | Median | AV High | AV Low | Ct High | Ct Low |
|---|---|---|---|---|---|---|---|
| 10 | 5 | 0.285548 | 0.27 | 0.436024 | 0.022465 | 264 | 151 |
| 25 | 20 | 0.255575 | 0.199658 | 0.411041 | 0.02904 | 255 | 175 |
| 50 | 45 | 0.272202 | 0.191097 | 0.421351 | 0.032497 | 270 | 168 |
| 100 | 95 | 0.276281 | 0.211038 | 0.434989 | 0.030236 | 262 | 169 |
| 250 | 245 | 0.262024 | 0.156977 | 0.428466 | 0.03452 | 231 | 169 |

Table 18. Performance of adaptive pruning with different levels of training and pruning points for 5 deep, 10 wide layered networks.

| Train | Prune | Mean | Median | AV High | AV Low | Ct High | Ct Low |
|---|---|---|---|---|---|---|---|
| 10 | 5 | 0.276749 | 0.233821 | 0.445981 | 0.024441 | 246 | 165 |
| 25 | 20 | 0.29337 | 0.2584 | 0.444981 | 0.029608 | 254 | 146 |
| 50 | 45 | 0.263532 | 0.193668 | 0.432604 | 0.030587 | 248 | 180 |
| 100 | 95 | 0.258895 | 0.166590 | 0.420279 | 0.031621 | 238 | 169 |
| 250 | 245 | 0.291801 | 0.24348 | 0.436872 | 0.036436 | 257 | 146 |

Tables 19 and 20 present data to characterize whether or not it makes a notable difference as to how far before the end of the training that the pruning is performed. No clear result is shown. Table 19 would tend to suggest that pruning early may have a limited positive impact on the mean error (however, its lowest median error comes when pruning at 90 epochs). Table 20 suggests the opposite, producing its lowest mean and median error when pruning and 95 epochs. Given this, there is no clear point at which pruning seems most valuable.

Table 19. Performance of adaptive pruning with different pruning points and 100 epochs of training for 5 deep, 5 wide layered networks.

| Prune | Mean | Median | AV High | AV Low | Ct High | Ct Low |
|---|---|---|---|---|---|---|
| 10 | 0.26621 | 0.202837 | 0.41188 | 0.029385 | 265 | 163 |
| 20 | 0.260715 | 0.212531 | 0.417192 | 0.029203 | 253 | 171 |
| 50 | 0.281493 | 0.243114 | 0.427361 | 0.025987 | 268 | 153 |
| 90 | 0.272063 | 0.199841 | 0.425719 | 0.031015 | 251 | 160 |
| 95 | 0.276281 | 0.211038 | 0.434989 | 0.030236 | 262 | 169 |

Table 20. Performance of adaptive pruning with different pruning points and 100 epochs of training for 5 deep, 10 wide layered networks.

| Prune | Mean | Median | AV High | AV Low | Ct High | Ct Low |
|---|---|---|---|---|---|---|
| 10 | 0.267054 | 0.209 | 0.426442 | 0.025954 | 239 | 158 |
| 20 | 0.263405 | 0.218168 | 0.410623 | 0.024966 | 264 | 163 |
| 50 | 0.267146 | 0.19725 | 0.422352 | 0.025814 | 269 | 173 |
| 90 | 0.281931 | 0.222104 | 0.436687 | 0.028781 | 265 | 162 |
| 95 | 0.258895 | 0.16659 | 0.420279 | 0.031621 | 238 | 169 |

The active filtering technique dramatically lowers the mean and median error by discarding poorly performing networks. Notably, while the performance can be assessed using the error level, this is not the metric used to determine whether to retain or discard candidate networks. As described in Section 4.3, this decision is based on whether rules are found to be impactful in decision-making. While not all high error level networks are discarded (approximately 25 to 40 remain, as shown in Tables 22 and 23), the error level of the high error networks is reduced (to 15% to 20%) along with their count. Thus, while producing approximately the same number of low error network completions, high error (demonstrably non-functional) networks are discarded, resulting in only effective networks (albeit, some with 15% to 20% error levels) being developed.

Table 21. Experimental hyperparameters.

| Model hyperparameters | | Experimental parameters | | Algorithm hyperparameters | | |
|---|---|---|---|---|---|---|
| Facts | Rules | Network | | Training epochs | Velocity | Training approach |
| Per condition | As required for connectivity | Layered – Config per condition<br>Pruning Type – Adaptive with active filtering<br>Prune Epoch – Per Condition | | 100 | 0.1 | Base |

Table 22. Performance of adaptive pruning with active filtering for networks with different pruning points and 100 epochs of training for 5 deep, 5 wide layered networks.

| Prune At | Mean | Median | AV High | AV Low | Ct High | Ct Low |
|---|---|---|---|---|---|---|
| 10 | 0.043206 | 0.018867 | 0.164882 | 0.021366 | 28 | 156 |
| 20 | 0.042516 | 0.021866 | 0.157141 | 0.025459 | 25 | 168 |
| 50 | 0.063217 | 0.027021 | 0.189302 | 0.028435 | 40 | 145 |
| 90 | 0.057504 | 0.02683 | 0.205 | 0.028005 | 28 | 140 |
| 95 | 0.055236 | 0.024877 | 0.174115 | 0.024144 | 34 | 130 |

Table 23. Performance of adaptive pruning with active filtering for networks with different pruning points and 100 epochs of training for 5 deep, 10 wide layered networks.

| Prune At | Mean | Median | AV High | AV Low | Ct High | Ct Low |
|---|---|---|---|---|---|---|
| 10 | 0.038749 | 0.012536 | 0.169848 | 0.020699 | 19 | 138 |
| 20 | 0.046544 | 0.021976 | 0.170937 | 0.024003 | 27 | 149 |
| 50 | 0.052569 | 0.02548 | 0.179036 | 0.026433 | 31 | 150 |
| 90 | 0.058832 | 0.029974 | 0.194895 | 0.02975 | 28 | 131 |
| 95 | 0.075653 | 0.035696 | 0.177859 | 0.026254 | 58 | 120 |

While the active filtering technique produces less than half the total number of completions, the completions that are produced have notably higher quality. Thus, adaptive filtering with active pruning appears to be a combined solution that can produce networks for use in many applications. Notably, the combined adaptive pruning and active filtering techniques outperform the base perfect network approach presented in Tables 2 and 4. Because active filtering is makes a decision based on the results of the adaptive pruning process, it is only usable with this technique. Creating a similar filtering mechanism for other pruning techniques is a potential topic for future work.

## 7. Comparison between same facts and random facts approaches

In prior work [8], several techniques were presented for training expert systems using gradient descent style techniques. Two of these techniques, which use multiple paths, are not readily adaptable to the pruning techniques presented herein. Their adaptation to use pruning remains a topic for future work.

This section assesses the efficacy of pruning for the second single-path technique, which utilizes random facts to simulate environments where training data varies to a great extent. Table 25 compares the two techniques when used with layer-based networks. Notably, the two perform similarly, with the random facts-based technique outperforming slightly in terms of both mean and median.

Table 24. Experimental hyperparameters.

| Model hyperparameters | | Experimental parameters | Algorithm hyperparameters | | |
|---|---|---|---|---|---|
| Facts | Rules | Network | Training epochs | Velocity | Training approach |
| 25 | 100 | Layered – 5W / 5D Pruning Type – None | 100 | 0.1 | Per condition |

Table 25. Comparison of train path – same facts and train path – random facts techniques.

| | Mean | Median | AV High | AV Low | Ct High | Ct Low |
|---|---|---|---|---|---|---|
| TPSF | 0.321855 | 0.295944 | 0.36373 | 0.049668 | 351 | 54 |
| TPRF | 0.302537 | 0.267088 | 0.356639 | 0.051079 | 330 | 71 |

Having established the base level of performance for the random facts technique, its performance with adaptive pruning is evaluated. A trend of increasing error with later pruning is shown in Table 27 (with relevant hyperparameters presented in Table 26); however, the level of difference between the performance at the different levels is marginal.

Table 26. Experimental hyperparameters.

| Model hyperparameters | | Experimental parameters | Algorithm hyperparameters | | |
|---|---|---|---|---|---|
| Facts | Rules | Network | Training epochs | Velocity | Training approach |
| 25 | 100 | Layered – 5W / 5D Pruning Type – Adaptive Prune Epoch – Per condition | 100 | 0.1 | TPRF |

Table 27. Train path – random facts training technique applied at different pruning points.

| Prn. Ep. | Mean | Median | AV High | AV Low | Ct High | Ct Low |
|---|---|---|---|---|---|---|
| 10 | 0.30258 | 0.266969 | 0.355566 | 0.052686 | 349 | 74 |
| 50 | 0.315099 | 0.271138 | 0.370539 | 0.051578 | 366 | 77 |
| 95 | 0.321273 | 0.289153 | 0.374948 | 0.053657 | 354 | 71 |

Table 29 shows that the adaptive pruning technique can have a similar level of reduction on the random facts-based technique (with relevant hyperparameters shown in Table 28). However, it is notable that the technique significantly reduces completions and does not increase the number of low-error completions as it did with the single facts-based techniques.

Table 28. Experimental hyperparameters.

| Model hyperparameters | | Experimental parameters | Algorithm hyperparameters | | |
|---|---|---|---|---|---|
| Facts | Rules | Network | Training epochs | Velocity | Training approach |
| 25 | 100 | Layered – 5W / 5D Pruning Type – Adaptive with active filtering Prune Epoch – Per condition | 100 | 0.1 | TPRF |

Table 29. Train path – random facts training technique with adaptive pruning and active filtering applied at different pruning points.

| Prn. Ep. | Mean | Median | AV High | AV Low | Ct High | Ct Low |
|---|---|---|---|---|---|---|
| 10 | 0.206765 | 0.169536 | 0.272628 | 0.0505 | 121 | 51 |
| 50 | 0.195202 | 0.162934 | 0.259848 | 0.052059 | 124 | 56 |
| 95 | 0.192776 | 0.166484 | 0.252192 | 0.051258 | 131 | 55 |

Thus, while adaptive pruning (and, by extension, active filtering) is useful for reducing error levels, the overall performance of the techniques is limited by the low completions produced. This may limit the efficacy of the technique for some applications where usable networks cannot be generated (or cannot be rapidly generated).

## 8. Manual processing of trained and pruned networks for defensibility

Once the automated initial training and pruning processes are complete, a key step for maintaining network defensibility must be performed: manual review and labeling. The initial training and pruning processes will (in most cases) reduce the number of facts utilized by the network. Some of these facts will be inputs, which are already labeled. Other facts will be intermediary conclusions within the network and, thus, not be initially labeled. Similarly, the number of rules interconnecting the input and intermediate facts will also be reduced.

These remaining rules and facts will have been shown by the training/pruning process to be effective at characterizing the phenomena, based on producing target outputs from input values provided. However, they may include the oversimplifications, non-causal connections and other maladies that have been discussed as being problematic in neural networks. The manual review process involves, thus, labeling all intermediate facts – either as functional facts that are used to facilitate grouping of facts between multiple system rules to form larger logical rules or as named meaningful values. Rules should, similarly, be identified as either functional or identified as representing a particular association between data elements. Any facts and rules that cannot be readily identified should be investigated, potentially improving knowledge of the phenomena that is being modeled – or to correct the model, if they are found to be erroneous.

Given the forgoing, the training/pruning process should be taken as an approach to speed up and potentially simplify network creation. Thus, it is the beginning of the process, not the end. Of course, this process may facilitate the creation of networks for phenomena that networks cannot be readily manually created for (just as neural networks have been implemented for phenomena that are not fully understood). Similarly, their use in solution identification should be the start of a manual understanding and review process – not a method for bypassing understanding on the way to system implementation. The manual review, labeling and correction step is absolutely integral to maintaining the key explainability and defensibility characteristics of the machine learning trained expert system technique. That being said, there may be other ways to further automate this process which can be explored as part of future work in network development automation.

## 9. Comparison to existing approaches

Several types of comparisons are relevant to the techniques presented herein. First, the accuracy of networks produced by the pruning techniques can be compared to the different approaches previously demonstrated using gradient descent trained expert systems and neural networks. Second, the procedures used by the proposed technique can be compared to other approaches for solving similar problems. Each is now discussed.

### *9.1. Comparison to Base Gradient Descent Trained Expert Systems Work*

Comparing the data presented in [8] and previously in this paper facilitates the understanding of the proposed approach's performance. In [8], four techniques were presented for using a priori known good networks and training to model phenomena. The average error for these techniques ranged from 5.2% to 8.3%. Two of these techniques, which train single pathways through the networks, were shown to be able to be further optimized, in [10], reducing their average error level down to 6.1% (93.9% accuracy) for the train path – same facts technique and 5.8% error (94.2% accuracy) for the train path – random facts technique. The error reduction techniques were not shown to be highly effective for the multiple path techniques. In all instances, networks based on the perfect network model (where analysis starts with a known network structure) were utilized. Overall, thus, the average performance of the currently known best techniques ranges between 5.2% and 6.1%. As was discussed in [8], this is well within the range of the performance of neural networks. Recent examples of neural networks performing with 97.7% [107], 85.9% [108], 98.3% [108] and 90.5% [109] accuracy were presented. Some neural network applications, of course, perform better and worse than these examples.

The data presented herein, for the adaptive pruned, active filtered layered networks falls within this same range of efficacy. In Tables 22 and 23, average error is shown to range from 4.2% to 6.3% for smaller networks and from 3.8% to 7.5% for larger networks. As these values are being produced by a technique setting (the point at which pruning is conducted), selection of the optimal one for each network size (which may also vary somewhat by application area) would be prudent. Thus, the best accuracy of the layer-based pruning approach is 95.8% for smaller networks and 96.2% for larger ones. This, interestingly, outperforms the use of the same (train path – same facts) training technique, in both [8] and [10], without the benefit of the a priori known network structure. The comparative performance of the different techniques is visually depicted in Figure 8.

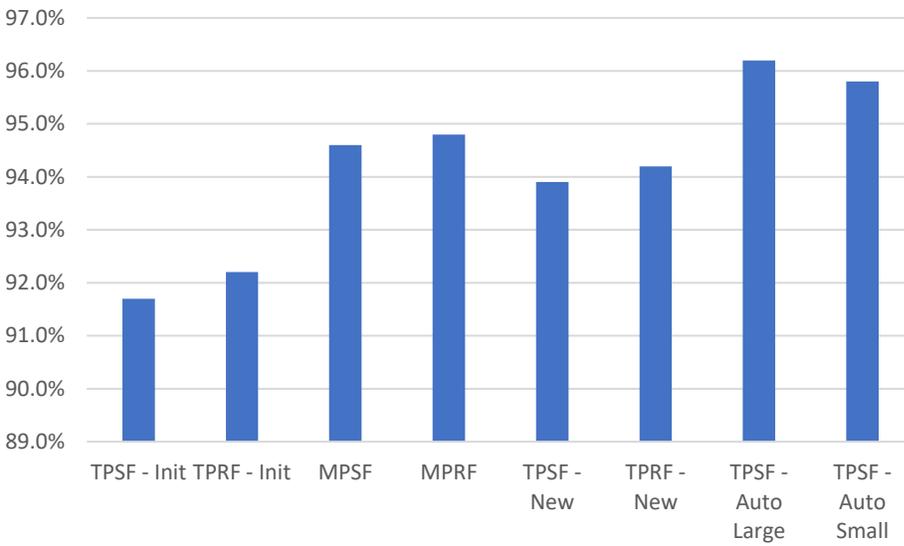

Figure 8. Comparison of the performance of different algorithm approaches.

## 9.2. Comparison of Procedures

The three different approaches also can be compared in terms of the procedures that they utilize. The prior work [8] on gradient descent trained expert systems required an a priori known model for the phenomena or that one be created. Neural networks, on the other hand, don't require a known network; however, in many cases they do require knowledge about characteristics of the phenomena that is being modeled to facilitate parameter selection – or a parameter optimization process [110–112] can be used.

The pruning-based gradient descent trained expert system approach presented in this paper doesn't require the a priori network knowledge that the original gradient descent trained expert system did; however, it does require some knowledge of the parameters of that network to determine an appropriate sized network for emulating it. In addition, the approach adds a manual labeling and review step.

Thus, in many cases, the time cost for implementing this model will fall in between that of implementing a neural network and the original gradient descent expert system. Both the neural network and pruning-based gradient descent trained expert system will require parameter determination. The model proposed herein also requires the labeling and review step, which is not needed for neural network implementation. This will, typically, be less involved than manual network creation (arguably, if it wasn't, it could simply be replaced by manual network creation in a particular case).

Of course, in the case where an existing system-usable model for the phenomena already existed, the approach proposed herein would underperform the one presented in [8] (which starts with the known-good network), in terms of implementation time. It would not be prudent to use the approach presented herein, in this case, unless trying to validate or further optimize the known model.

Both the approach presented in [8] and the one presented herein provide the key benefit of defensibility. In both cases, the system is not able to – once an initial network has been developed by a human or automatically developed and human reviewed – deviate from the known-reliable rules and meaningful facts of the expert system during optimization. In both cases, though, the system can further optimize rule weightings to enhance system performance over time.

The benefits of the different approaches discussed in this section are visually summarized in Table 30.

Table 30. Comparison of benefits of different approaches.

|  | Manual GDTES | Automated GDTES | Neural Network |
|---|---|---|---|
| Learn / optimize from training data | ✓ | ✓ | ✓ |
| A prior model / model creation | ✓ | ✗ | ✗ |
| Knowledge of phenomena / parameters | ✗ | ✓ | ✓ |
| Post-network creation manual review | ✗ | ✓ | ✗ |
| Performance with existing model | High | Lower | Lower |
| Creation Time | Highest | Middle | Lowest |
| Defensibility (cannot learn invalid relationships) | ✓ | ✓* | ✗ |

\* Benefit occurs based on manual review step

## 10. Conclusions and future work

Fundamentally, the contribution of this paper has been to demonstrate the potential to automate the creation of gradient descent-trained expert system networks. It has shown the efficacy of using pruning to more rapidly develop a defensible and trainable rule-fact network expert system for various phenomena. The data has shown that, using adaptive pruning with active filtering, mean error rates as low as 3.9% (1.2% median) can be obtained. It has proposed the use of a larger expert system rule-fact network with training and pruning to produce an optimized (and smaller) gradient descent trained expert system network for various phenomena across different application areas.

This approach aims to bring the time costs of system development for gradient descent trained expert systems closer to those of neural networks (though, in most cases, not reaching that of neural networks) while still providing the defensibility (known understandable and causal networks) and explainability benefits of the original gradient descent trained expert systems. It also aims to solve the issue of the difficulty of model development for application areas that have not been extensively studied, by creating a template model for human review and refinement.

The data presented herein shows that the combination of an adaptive pruning technique with active filtering enhances the performance of layered networks to a point where they produce results similar to (and slightly superior to) the 'perfect' network based systems presented in [8]. The data also showed that this approach produces results that are within the range of the performance of neural network techniques, for many application areas.

The current work has also identified a number of key areas of needed future work. Additional work is need to focus on whether a similar pruning technique can be developed for multi-path training techniques (which outperformed the single path techniques in [8] and could prospectively outperform the results presented herein, if a suitable pruning technique was developed for them). Similarly, more work is needed to optimize the pruning techniques for the single path – random facts training technique, discussed herein.

**Acknowledgments**

Thanks are given to two anonymous reviewers whose suggestions improved this paper. The visualizations shown in Figures 3(a) and 3(b) were produced using GraphViz based upon simulation system network output data.